\documentclass{article}

\usepackage{arxiv}

\usepackage[utf8]{inputenc} 
\usepackage[T1]{fontenc}    
\usepackage{hyperref}       
\usepackage{url}            
\usepackage{booktabs}       
\usepackage{amsfonts}       
\usepackage{nicefrac}       
\usepackage{microtype}      
\usepackage{lipsum}
\usepackage{graphicx}
\graphicspath{ {./images/} }
\usepackage{graphicx}
\usepackage{natbib}
\usepackage{adjustbox}
\usepackage{float}
\usepackage{amsmath}
\usepackage{amssymb}
\usepackage{multirow}
\usepackage{multicol}
\usepackage{xcolor}
\usepackage{doi}
\usepackage{pifont}
\usepackage{threeparttable}
\usepackage{tabularx}

\title{Efficient Chest X-ray Representation Learning via Semantic-Partitioned Contrastive Learning}

\author{{\hspace{1mm}Wangyu Feng}\thanks{Wangyu Feng and Shawn Young contributed equally to this work and conducted it during an internship at Shenzhen University of Advanced Technology.} \\
	\textsuperscript{1}Shenzhen University of Advanced Technology\\
	Shenzhen, China \\
	\And
	\hspace{1mm}Shawn Young\textsuperscript{*} \\
	\textsuperscript{1}Shenzhen University of Advanced Technology\\
	Shenzhen, China \\
    \And
	\hspace{1mm}Lijian Xu\thanks{Corresponding author.} \\
	Shenzhen University of Advanced Technology \\
	Shenzhen, China \\
	\texttt{xulijian@suat-sz.edu.cn} \\
}

\begin{document}
\maketitle
\begin{abstract}
Self-supervised learning (SSL) has emerged as a powerful paradigm for Chest X-ray (CXR) analysis under limited annotations. 
Yet, existing SSL strategies remain suboptimal for medical imaging. 
Masked image modeling allocates substantial computation to reconstructing high-frequency background details with limited diagnostic value. 
Contrastive learning, on the other hand, often depends on aggressive augmentations that risk altering clinically meaningful structures.
We introduce \textbf{Semantic-Partitioned Contrastive Learning (S-PCL)}, an efficient pre-training framework tailored for CXR representation learning. 
Instead of reconstructing pixels or relying on heavy augmentations, S-PCL randomly partitions patch tokens from a single CXR into two non-overlapping semantic subsets. 
Each subset provides a complementary but incomplete view. 
The encoder must maximize agreement between these partitions, implicitly inferring global anatomical layout and local pathological cues from partial evidence. 
This semantic partitioning forms an internal bottleneck that enforces long-range dependency modeling and structural coherence.
S-PCL eliminates the need for hand-crafted augmentations, auxiliary decoders, and momentum encoders. 
The resulting architecture is streamlined, computationally efficient, and easy to scale. 
Extensive experiments on large-scale CXR benchmarks, including ChestX-ray14, CheXpert, RSNA Pneumonia and SIIM-ACR Pneumothorax, show that S-PCL achieves competitive performance while attaining the lowest GFLOPs and superior accuracy among existing SSL approaches.
The code is available at \url{https://anonymous.4open.science/r/SPCL-C621}. 
\end{abstract}


\section{Introduction}

Self-supervised learning has become a central paradigm in medical image analysis by enabling representation learning from large-scale unlabeled data~\cite{chen2019self,young2026scalar}.
Its effectiveness has been validated across classification, segmentation, and cross-domain transfer scenarios~\cite{zhou2020comparing,yang2023geometry}, with subsequent studies demonstrating robustness and scalability in large medical corpora~\cite{liu2023m}.

Existing approaches largely fall into two categories: purely visual pretraining and vision language alignment.
Within visual pretraining, image-level contrastive learning promotes instance discrimination and shows strong transferability~\cite{chen2020simple,yang2025one}.
Masked image modeling instead learns contextual dependencies through reconstruction objectives~\cite{xiao2023delving}, and has been extended to multi-scale and volumetric medical data~\cite{chen2022multi,xu2024foundation,xu2024medvilam}.
Large-scale pretraining further improves performance across clinical benchmarks~\cite{wu2025large,yang2024segmentation}.
Multimodal learning incorporates radiology reports to enhance semantic alignment~\cite{Zhou_2022,huang2024enhancing,xu2023learning}.
Contrastive cross-modal strategies strengthen image–text correspondence~\cite{zhang2022contrastive,huang2021gloria}, while structured knowledge integration further refines chest X-ray representation~\cite{boecking2022making,wu2023medklip,wang2022multi}.
Recently, efficient token compression has been widely studied to reduce computational overhead while preserving semantic content~\cite{young2026fewer,he2026autoselect,chen2026tc,gao2026zerosense,wu2026multimodal,chen2026multimodal}.

Despite these advances, current paradigms remain suboptimal for chest X-rays.
Reconstruction-based objectives emphasize pixel fidelity~\cite{xiao2023delving,zhou2023advancing}, often biasing models toward high-frequency textures.
Contrastive frameworks rely on strong augmentations~\cite{chen2020simple}, which may distort subtle anatomical cues.
Vision language approaches partially alleviate semantic ambiguity~\cite{Zhou_2022,wu2023medklip}, yet still depend on the stability of the underlying visual encoder.
Taken together, existing strategies either optimize low-level reconstruction, enforce potentially unsafe invariances, or depend on auxiliary textual supervision.
They do not explicitly exploit the structural property of chest X-rays, where diagnostic information is spatially sparse yet globally organized.
A self-supervised objective that captures holistic anatomical relationships without reconstruction overhead or augmentation-induced distortion therefore remains desirable.

To bridge this gap, we introduce S-PCL (Semantic-Partitioned Contrastive Learning), a simple yet effective self-supervised framework tailored for chest X-ray representation learning.
Instead of reconstructing masked regions or relying on hand-crafted augmentations, S-PCL constructs two complementary views by randomly partitioning image patches into non-overlapping subsets within a single image.
By maximizing agreement between these partitioned views, the encoder is encouraged to infer missing contextual information and capture coherent global thoracic structure.
Such a design promotes the modeling of clinically meaningful anatomical relationships, for example, the spatial dependency between lung fields and ribs, while avoiding pixel-level redundancy and unnecessary semantic distortion.
Our main contributions are summarized as follows:

1) We introduce S-PCL, a streamlined pre-training framework that integrates the efficiency of partition-based modeling with the discriminative power of contrastive learning, avoiding reconstruction overhead and augmentation-induced distortion.

2) We show that contrasting non-overlapping partitions enables efficient learning of high-level diagnostic representations without auxiliary components such as momentum encoders or complex decoders.

3) Extensive experiments on large-scale CXR datasets demonstrate state-of-the-art downstream performance and favorable scaling efficiency.

\begin{figure}[!t]
    \centering
    \includegraphics[width=\linewidth, height=\textheight, keepaspectratio]{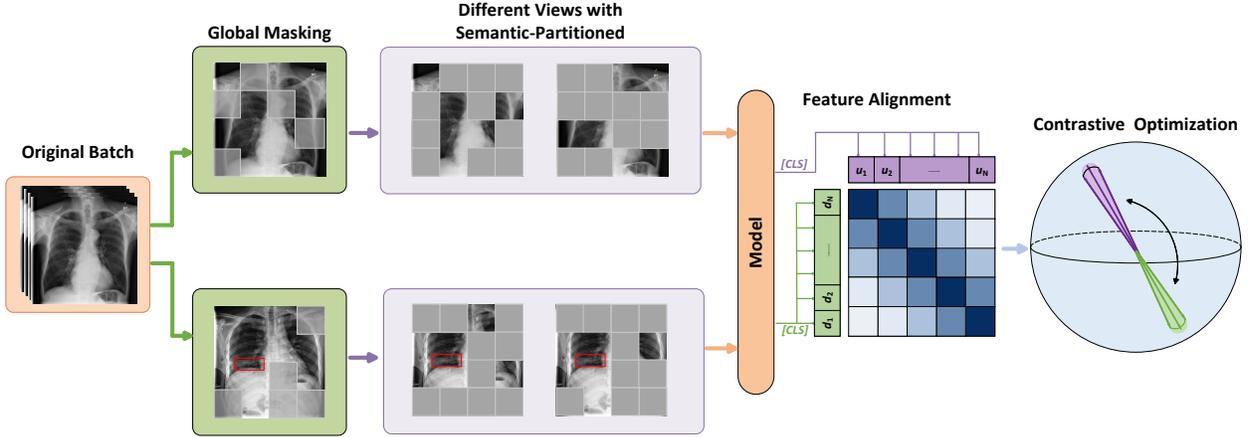}
    \caption{Our occluded image contrastive learning: Through non-overlapping occluding, distinct tokens within an image are categorized as intraclass, while across-image tokens within a batch are viewed as interclass. }
    \label{fig:fig1}
\end{figure}

\section{Methodology}

\subsection{Method Overview}

Avoiding aggressive hand-crafted augmentations, S-PCL executes its self-supervised task in three steps (see Figure~\ref{fig:fig1}). 1) tokenizing and randomly masking the input radiograph; 2) splitting the visible tokens into disjoint subsets to create parallel contrastive views; and 3) passing these views through a shared Vision Transformer to optimize a T-distributed spherical contrastive loss.

To formalize this pipeline, we first detail the initial tokenization stage. Given an input CXR image $\mathbf{x} \in \mathbb{R}^{H \times W \times 1}$, we follow the standard Vision Transformer (ViT) protocol by reshaping it into a sequence of 2D patches $\mathbf{x}_p \in \mathbb{R}^{N \times P^2}$, where $(H, W)$ denotes the original image resolution, $P$ represents the patch size, and $N = HW/P^2$ indicates the total number of patches. These patches are subsequently mapped via a linear projection into $D$-dimensional embeddings and combined with fixed sinusoidal positional embeddings $\mathbf{p} \in \mathbb{R}^{N \times D}$ to preserve the thoracic anatomical layout. Consequently, this yields a token sequence $\mathbf{E} = \text{Linear}(\mathbf{x}_p) + \mathbf{p} \in \mathbb{R}^{N \times D}$ encapsulating the fine-grained semantics of the image before the masking operation. This unmasked sequence $\mathbf{E}$ establishes the foundational representation from which we derive our computationally efficient contrastive views, as detailed in the subsequent masking strategy.

\subsection{Semantic-Partitioned Masking Strategy}

Medical images inherently exhibit significant conceptual redundancy. While various patches within a single radiograph display distinct visual characteristics, they are all fundamentally interconnected with the overall clinical meaning of the image. Instead of reconstructing masked pixels as in standard Masked Image Modeling (MIM), our strategy utilizes random sampling without replacement following a uniform distribution to initially mask the image. 

Given a global masking ratio $r$ (e.g., $30\%$), the remaining $n = \lfloor (1-r)N \rfloor$ visible patches are retained to capture the comprehensive anatomical structure. To construct the contrastive views, these visible patches are randomly partitioned into two non-overlapping groups $\mathcal{V}_{1}, \mathcal{V}_{2} \in \mathbb{R}^{\frac{n}{2} \times D}$. This random, non-overlapping division ensures that the central positions of the visible patches remain statistically consistent across both groups while actively avoiding potential spatial biases. 

Crucially, this strategy creates a deliberate dual-ratio effect. While the global masking ratio remains low to capture overall relationships, the disjoint partitioning means each contrastive branch effectively experiences a much higher masking ratio (e.g., $65\%$). During each forward pass, the model observes only a severely restricted portion of the visible tokens. This challenging scenario forces the model to focus on localized pathological features rather than relying on simple patterns or redundant information, thereby creating diverse views with substantial fine-grained semantic differences at the concept level.

\subsection{Efficient Contrastive Learning}

Following the generation of the disjoint subsets $\mathcal{V}_{1}$ and $\mathcal{V}_{2}$, we prepend a learnable \texttt{[CLS]} token to each sequence. Both subsets are then fed independently into a shared Vision Transformer encoder, denoted as $f_{\theta}(\cdot)$. Let $\mathbf{z}_{1} = f_{\theta}(\mathcal{V}_{1})$ and $\mathbf{z}_{2} = f_{\theta}(\mathcal{V}_{2})$ represent the high-level embeddings extracted from the respective \texttt{[CLS]} tokens, as these tokens aggregate the high-level semantic information of each partitioned view. By processing these non-overlapping views independently, the framework establishes a strict internal information bottleneck. The encoder's self-attention mechanism is heavily constrained, forcing it to aggregate long-range dependencies across the thoracic cavity and implicitly infer both the global anatomical layout and localized pathological anomalies from severely partial visual evidence.

To optimize the encoder, we apply a contrastive learning objective that maximizes the representational agreement between the paired semantic partitions. For a given mini-batch containing $N$ CXR images, the masking and encoding process yields $2N$ representations. The embeddings $\mathbf{z}_{1}$ and $\mathbf{z}_{2}$ originating from the same radiograph constitute a positive pair, while the remaining $2(N-1)$ representations within the same batch serve as negative samples. In terms of similarity computation, we introduce the T-distributed spherical (T-SP) metric to significantly promote the intraclass compactness and interclass separability of features. Given the cosine distance $\cos(\mathbf{z}_{1}, \mathbf{z}_{2})$ between the two normalized feature vectors, the T-SP similarity is defined as:
\begin{equation}
    sim_{tsp}(\mathbf{z}_{1}, \mathbf{z}_{2}) = 0.5 \times \frac{1 + \cos(\mathbf{z}_{1}, \mathbf{z}_{2})}{1 + (1 - \cos(\mathbf{z}_{1}, \mathbf{z}_{2})) \times \kappa}
\end{equation}
where $\kappa > 0$ denotes the concentration hyperpa
rameter of the T-SP metric. We concurrently maximize the similarity of positive pairs while minimizing the similarity of negative examples to drive the network. The loss function for a positive pair is formulated as:
\begin{equation}
    \mathcal{L} = - \log \frac{\exp(sim_{tsp}(\mathbf{z}_{1}, \mathbf{z}_{2}) \times \tau)}{\sum_{j=1}^{2N} \mathbb{1}_{[j \neq 1]} \exp(sim_{tsp}(\mathbf{z}_{1}, \mathbf{z}_{j}) \times \tau)}
\end{equation}
where $\tau$ is a trainable temperature parameter used to effectively scale the different samples, and $\mathbb{1}$ is an indicator function evaluating to $1$ when the condition is met. This objective effectively enhances the intraclass conceptual compactness within a single radiograph and the interclass semantic separability across different patients in the mini-batch.

This semantic-partitioned contrastive design (S-PCL) is highly efficient. By defining the pre-training objective purely in latent space via representation alignment, it avoids the pixel-level reconstruction overhead. In contrast, standard MIM often overemphasizes high-frequency details and local features, which can diverge from the goal of learning high-level semantic concepts and reduce pre-training efficiency. Moreover, unlike many contrastive frameworks that require heavy pre-processing or auxiliary networks to create views, S-PCL uses the ViT \texttt{[CLS]} token directly, without momentum encoders, projection MLP heads, or decoders. This streamlined design reduces computation and memory, achieving the lowest GFLOPs among comparable SSL methods and scaling well to large clinical datasets.

\section{Experiments}
\subsection{Datasets and Implementation Details}
\noindent\textbf{Datasets.} 
We use \textbf{{MIMIC-CXR-JPG\cite{johnson2019mimic}}} for self-supervised pre-training. Following a 7:2:1 train-validation-test split across all downstream tasks, we evaluate our model on the following benchmarks:
1) \textbf{ChestX-ray14\cite{wang2017chestx}}, containing 112,120 radiographs with 14 pathology labels for multi-label classification.
2) \textbf{CheXpert \cite{irvin2019chexpert}}, comprising 224,316 images with 14 observations for multi-label classification.
3) \textbf{RSNA Pneumonia \cite{shih2019augmenting}}, containing approximately 30,000 images annotated for pulmonary opacity.
4) \textbf{SIIM-ACR Pneumothorax\cite{kaggle-siim}}, consisting of over 12,000 radiographs with pixel-level masks for semantic segmentation.

\begin{table}[t]
    \centering
    \caption{\textbf{Comparison of our method with specialist approaches for disease classification.} Performance is measured using macro‑averaged AUC. The labeling ratio X\% indicates the proportion of the fully annotated training set used for supervised fine‑tuning. Best results are highlighted in bold. }
    \label{tab:comparison_sota}
    \resizebox{\textwidth}{!}{
        \begin{tabular}{l c c ccc ccc ccc}
            \toprule
            \multirow{2}{*}{Method} & \multirow{2}{*}{GFLOPs} & \multirow{2}{*}{\shortstack{GPU\\Hrs}} & \multicolumn{3}{c}{ChestX-ray14} & \multicolumn{3}{c}{CheXpert} & \multicolumn{3}{c}{RSNA Pneu.} \\
            \cmidrule(lr){4-6} \cmidrule(lr){7-9} \cmidrule(lr){10-12}
             & & & 1\% & 10\% & 100\% & 1\% & 10\% & 100\% & 1\% & 10\% & 100\% \\
            \midrule
            \multicolumn{12}{l}{\textit{\color{gray} Training with clinical reports}} \\
            \midrule
            M3AE~\cite{chen2022multi} & 34.4 & 710 & - & - & - & 86.2 & 87.3 & 87.9 & 89.0 & 90.8 & 92.3 \\
            REFERS~\cite{Zhou_2022} & 54.8 & - & 76.7 & 80.9 & 84.7 & 87.2 & 88.1 & 88.2 & 89.4 & 91.6 & 92.7 \\
            MRM~\cite{zhou2023advancing} & 13.1 & 800 & 79.4 & 84.0 & 85.9 & 88.5 & 88.5 & 88.7 & 91.3 & 92.7 & 93.3 \\
            \midrule
            \multicolumn{12}{l}{\textit{\color{gray} Training without clinical reports}} \\
            \midrule
            SimCLR~\cite{chen2020simple} & 8.2 & 830 & - & - & - & - & - & - & 70.1 & 80.2 & 84.9 \\
            Medical MAE~\cite{xiao2023delving} & 20.3 & 1200 & - & - & 82.3 & - & - & 89.2 & - & - & - \\
            CheXWorld~\cite{yue2025chexworld} & 67.4 & - & - & - & 83.5 & - & - & \textbf{89.6} & - & - & 75.0 \\
            \textbf{S-PCL (Ours)} & \textbf{6.1} & \textbf{540} & \textbf{78.2} & \textbf{82.1} & \textbf{84.1} & \textbf{86.7} & \textbf{88.4} & 89.1 & \textbf{86.6} & \textbf{89.2} & \textbf{91.2} \\
            \bottomrule
        \end{tabular}
    }
\end{table}

  \begin{figure}[!t]
    \centering
    \includegraphics[width=0.65\linewidth]{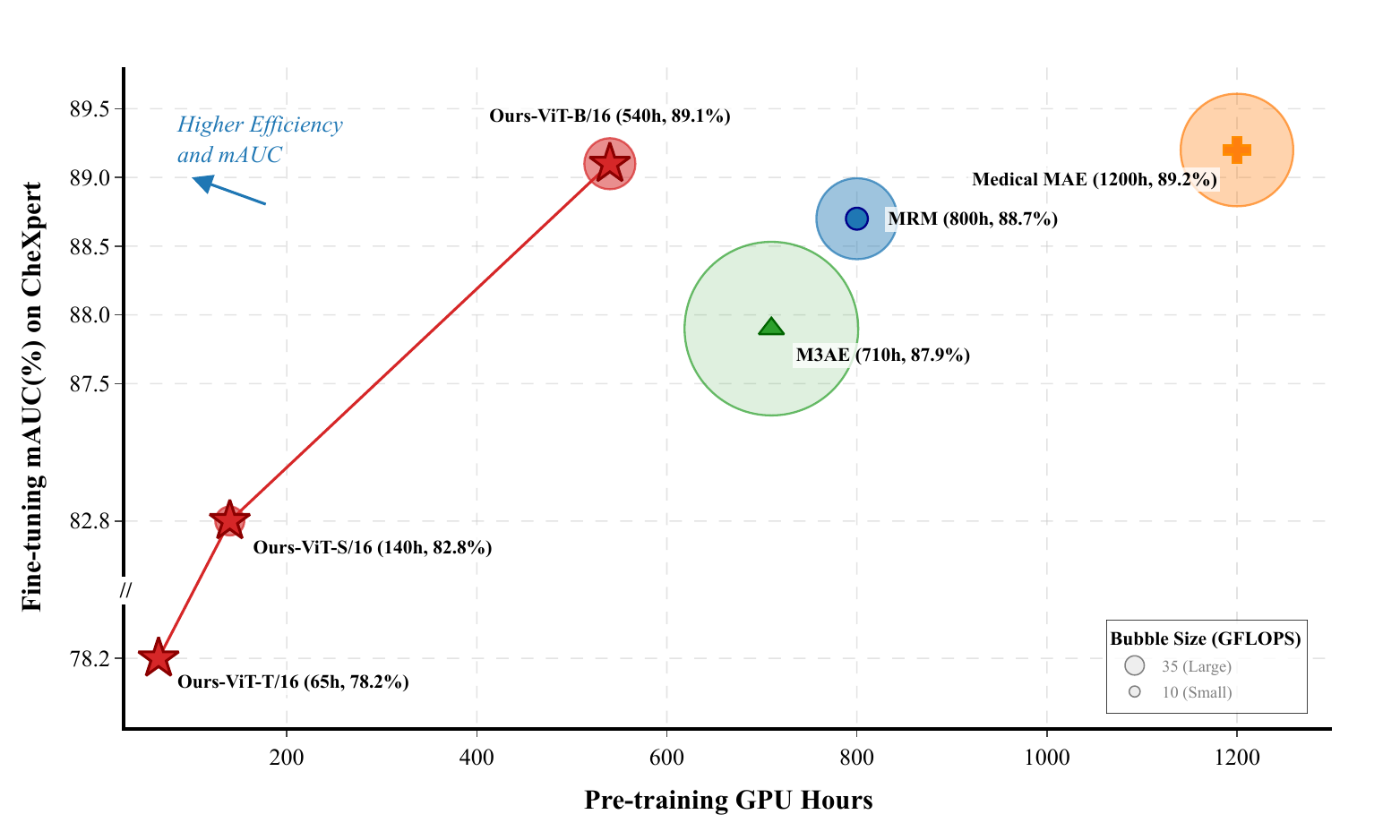}
    \caption{Efficiency and scaling comparison of MRM\cite{zhou2023advancing}, Medical MAE\cite{xiao2023delving},  M3AE~\cite{chen2022multi}, and our method on CheXpert fine-tuning, measured by mAUC (\%).}
    \label{fig:fig2}
  \end{figure}

\noindent\textbf{Pre-training and Fine-tuning Settings.} We employed a ViT-Base backbone using $16 \times 16$ patches, which encodes input images of size $512 \times 512$ into a sequence of patch tokens. During each forward pass, a random subset of patch tokens is masked at a ratio of 0.6, and the remaining visible tokens are split into two non-overlapping views that are independently encoded by the shared ViT encoder. A contrastive loss is then computed between the \texttt{[CLS]} token representations of the two branches. We used a mini-batch size of 600 per GPU with gradient accumulation over 2 iterations. Furthermore, we follow the fine-tuning settings of Medical MAE\cite{xiao2023delving}. To ensure consistency, all training and inference times are measured using a single NVIDIA RTX A6000 GPU.

\subsection{Comparison with State-of-the-art Methods }
Figure~\ref{fig:fig2} shows that among the compared methods, S-PCL achieves the lowest pre-training cost (540 GPU hours) while maintaining competitive fine-tuning performance (89.1\% mAUC). In contrast, Medical MAE requires the most pre-training resources (1200 GPU hours) for a marginal performance gain (89.2\% mAUC), and MRM offers a balanced trade-off with 800 GPU hours and 88.7\% mAUC. This suggests that S-PCL is more efficient in pre-training without significant sacrifice in downstream accuracy. Furthermore, employing ViT-S/16 as the backbone, our S-PCL reaches 82.8\% mAUC on the CheXpert dataset, requiring only 140 GPU hours for pre-training. When using the lighter ViT-T/16 backbone, S-PCL still achieves 78.2\% mAUC with only 65 GPU hours of pre-training, further demonstrating its favorable efficiency-performance trade-off across different model scales.

Table~\ref{tab:comparison_sota} shows that S-PCL achieves competitive performance across all three chest X-ray datasets under different fine-tuning ratios. On ChestX-ray14, S-PCL attains 78.2\%, 82.1\%, and 84.1\% AUC with 1\%, 10\%, and 100\% of the training data, respectively. On CheXpert, it reaches 86.7\%, 88.4\%, and 89.1\% AUC, while on RSNA Pneumonia, it obtains 86.6\%, 89.2\%, and 91.2\% AUC.

\begin{table*}[t]
\centering
\footnotesize
\caption{\textbf{Performance comparison on the CheXpert benchmark}. We report the Area Under the Curve (AUC) for five clinical observations and their mean (mAUC). Best results are highlighted in bold. }
\label{tab:sota_chexpert}
\resizebox{\textwidth}{!}{
\begin{tabular}{llcccccc}
\toprule
Method & Backbone & Atelectasis & Cardiomegaly & Consolidation & Edema & Effusion & mAUC (\%) \\
\midrule
Label-Assemble~\cite{kang2023label}& DN121 & 82.1 & 85.9 & \textbf{94.4} & 89.2 & 93.6 & 89.0 \\
Medical MAE~\cite{xiao2023delving} & ViT-B/16 & \textbf{82.7} & 83.5 & 92.5 & 93.8 & 94.1 & \textbf{89.3} \\
\midrule
\textbf{S-PCL (Ours)} & ViT-B/16 & 80.3 & \textbf{95.4} & 80.4 & \textbf{94.1} & \textbf{95.6} & 89.1\\
\bottomrule

\end{tabular}
}
\end{table*}

\begin{table}[!b]
    \caption{\textbf{Performance comparison on the ChestX-ray14 benchmark.} We report the AUC scores across all 14 categories. The best results are in bold and the second-best results are underlined. }
    \centering
    \label{tab:sota_cxr14}
    \resizebox{\textwidth}{!}{
        \begin{tabular}{l *{15}{c}}
        \toprule
         Model & \rotatebox{90}{Mean} & \rotatebox{90}{Atelectasis} & \rotatebox{90}{Cardiomegaly} & \rotatebox{90}{Effusion} & \rotatebox{90}{Infiltration} & \rotatebox{90}{Mass} & \rotatebox{90}{Nodule} & \rotatebox{90}{Pneumonia} & \rotatebox{90}{Pneumothorax} & \rotatebox{90}{Consolidation} & \rotatebox{90}{Edema} & \rotatebox{90}{Emphysema} & \rotatebox{90}{Fibrosis} & \rotatebox{90}{Pleural Thicken} & \rotatebox{90}{Hernia} \\
        \midrule
         ConVIRT\cite{zhang2022contrastive} & 56.0 & 45.1 & 44.3 & 63.2 & 65.1 & 61.6 & 57.2 & 63.6 & 54.1 & 63.7 & 70.2 & 41.9 & 47.4 & 56.0 & 51.2 \\
         GLoRIA\cite{huang2021gloria}  &81.8  &\underline{82.6}  &83.3  &86.0  &66.4  &\underline{81.8} &73.5  &71.0  &84.5  &\underline{81.3}  &\underline{89.8} &93.1 &78.9  &76.1  & \textbf{97.5} \\
         MRM~\cite{zhou2023advancing} &\textbf{85.9} &\textbf{84.2} &\textbf{93.0} &\underline{89.6} &\underline{71.8} &\textbf{88.2} &\textbf{78.5} & \underline{77.3} &\underline{90.2} &\textbf{82.2} &\textbf{91.0} &\textbf{94.3} & \textbf{86.7} &\textbf{81.4} &\underline{94.4}\\ 
         \midrule
         \textbf{S-PCL (Ours)}          & \underline{84.1} &82.3 &\underline{87.7}  &\textbf{91.4}  &\textbf{75.7}  &80.8 &\underline{75.0} &\textbf{77.3} &\textbf{92.5}  &79.0 &89.4 &\underline{93.5} &\underline{82.6}  &\underline{80.1} &90.3 \\
        \bottomrule
        \end{tabular}
    }
\end{table}

Table~\ref{tab:sota_chexpert} provides a detailed per-disease comparison on the CheXpert benchmark. S-PCL with a ViT-B/16 backbone achieves 89.1\% mean AUC, which is competitive with the best-performing methods. Notably, S-PCL obtains the highest AUC scores for Cardiomegaly (95.4\%), Edema (94.1\%), and Effusion (95.6\%), outperforming all other approaches including Medical MAE variants. This demonstrates that S-PCL is particularly effective at learning representations for these specific thoracic conditions.

Table~\ref{tab:sota_cxr14} further demonstrates the model's diagnostic capabilities across 14 diverse thoracic pathologies on ChestX-ray14 dataset. Our approach exhibits superior discriminative power in identifying specific localized conditions. Most notably, S-PCL excels in detecting complex anomalies such as Effusion (91.4\%) and Pneumothorax (92.5\%). These substantial gains indicate that forcing the encoder to infer global context from partitioned views effectively enhances its sensitivity to subtle, fine-grained pathological features compared to conventional contrastive frameworks.

\begin{table*}[t]
\centering
\small
\begin{threeparttable}
\caption{\textbf{Results of semantic segmentation and object detection.} Best results are highlighted in bold. Each model is fine-tuned with 1\%, 10\%, 100\% training data.}
\label{tab:seg_det_results}
\begin{tabularx}{\textwidth}{l | >{\centering\arraybackslash}X >{\centering\arraybackslash}X >{\centering\arraybackslash}X | >{\centering\arraybackslash}X >{\centering\arraybackslash}X >{\centering\arraybackslash}X}
\hline
\multirow{3}{*}{Methods} & \multicolumn{3}{c|}{\textbf{Semantic Segmentation}} & \multicolumn{3}{c}{\textbf{Object Detection}} \\
\cline{2-7}
& \multicolumn{3}{c|}{SIIM-ACR Pneumothorax} & \multicolumn{3}{c}{RSNA Pneumonia} \\
\cline{2-7}
& 1\% & 10\% & 100\% & 1\% & 10\% & 100\% \\
\hline
GLoRIA~\cite{huang2021gloria} & 37.4 & 57.1 & 64.0 & 11.6 & 16.1 & 24.8 \\
MGCA~\cite{wang2022multi}    & 49.7 & 59.3 & 64.2 & 12.9 & 16.8 & 24.9 \\
MedKLIP~\cite{wu2023medklip} & \textbf{50.2} & 60.8 & 63.9 & 8.9 & 16.3 & 24.5 \\
\hline
\textbf{S-PCL (Ours)} & 38.8 & \textbf{61.7} & \textbf{65.1} & \textbf{13.2} & \textbf{16.9} & \textbf{25.6} \\
\hline
\end{tabularx}
\end{threeparttable}
\end{table*}

Table~\ref{tab:seg_det_results} further demonstrates that S-PCL achieves competitive performance on semantic segmentation and object detection benchmark. The improvements are particularly evident in the 10\% and 100\% supervision settings on semantic segmentation benchmark and all settings on object detection becnmark, indicating strong dense prediction capability compared to prior vision–language pretraining methods.

\subsection{Feature Interpretability}

\begin{figure}[t]
\centering
\includegraphics[width=0.45\linewidth]{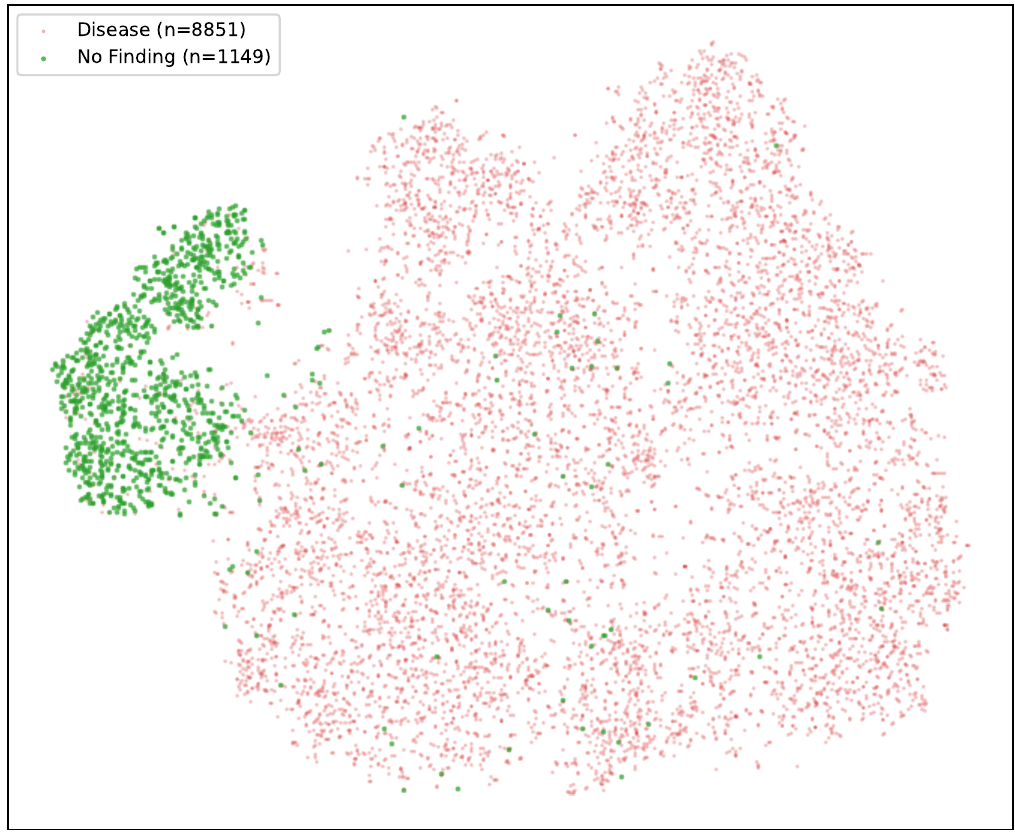} 
\caption{t-SNE visualization of the learned global representations on CheXpert benchmark.}
\label{fig:fig3}
\end{figure}

To interpret the learned representations, we visualize the high-dimensional features of 10,000 test images using t-SNE. We extract the global representations for 8,851 scans with diseases and 1,149 normal scans. Figure~\ref{fig:fig3} reveals a clear separation between pathological and normal radiographs. This demonstrates that by enforcing agreement between incomplete semantic partitions, S-PCL implicitly discovers highly discriminative and interpretable clinical concepts without relying on any explicit annotations.

\section{Conclusion}

We introduce S-PCL, a pre-training paradigm for CXRs that eliminates pixel-level reconstruction and risky augmentations by contrasting semantic-partitioned views. By exploiting inherent medical imaging redundancy, S-PCL efficiently learns robust representations from partial observations. This significantly reduces computational overhead while outperforming SOTA baselines, paving the way for scalable learning in high-resolution medical foundation models.

\bibliographystyle{unsrt}  
\bibliography{references}

@misc{kaggle-siim,
  title = {Society for Imaging Informatics in Medicine: Siim-acr pneumothorax segmentation.},
  howpublished = {\url{https://www.kaggle.com/c/siim-acr-pneumothorax-segmentation}},
  note = {2019}
}

@article{johnson2019mimic,
  title={Mimic-cxr database},
  author={Johnson, AEWP and Pollard, Tom and Mark, Roger and Berkowitz, Seth and Horng, Steven},
  journal={PhysioNet10},
  volume={13026},
  pages={C2JT1Q},
  year={2019}
}

@inproceedings{wu2023medklip,
  title={Medklip: Medical knowledge enhanced language-image pre-training for x-ray diagnosis},
  author={Wu, Chaoyi and Zhang, Xiaoman and Zhang, Ya and Wang, Yanfeng and Xie, Weidi},
  booktitle={ICCV},
  pages={21372--21383},
  year={2023}
}

@inproceedings{xiao2023delving,
  title={Delving into masked autoencoders for multi-label thorax disease classification},
  author={Xiao, Junfei and Bai, Yutong and Yuille, Alan and Zhou, Zongwei},
  booktitle={Proceedings of the IEEE/CVF Winter Conference on Applications of Computer Vision},
  pages={3588--3600},
  year={2023}
}

@inproceedings{huang2021gloria,
  title={GLoRIA: A Multimodal Global-Local Representation Learning Framework for Label-Efficient Medical Image Recognition},
  author={Huang, Shih-Cheng and Shen, Liyue and Lungren, Matthew P and Yeung, Serena},
  booktitle={Proceedings of the IEEE/CVF International Conference on Computer Vision},
  pages={3942--3951},
  year={2021}
}

@inproceedings{zhou2023advancing,
title={Advancing Radiograph Representation Learning with Masked Record Modeling},
author={Hong-Yu Zhou and Chenyu Lian and Liansheng Wang and Yizhou Yu},
booktitle={The Eleventh International Conference on Learning Representations },
year={2023}}

@article{Zhou_2022,
   title={Generalized radiograph representation learning via cross-supervision between images and free-text radiology reports},
   volume={4},
   ISSN={2522-5839},
   number={1},
   journal={Nature Machine Intelligence},
   publisher={Springer Science and Business Media LLC},
   author={Zhou, Hong-Yu and Chen, Xiaoyu and Zhang, Yinghao and Luo, Ruibang and Wang, Liansheng and Yu, Yizhou},
   year={2022},
   month=jan, pages={32–40} 
}

@inproceedings{zhang2022contrastive,
  title={Contrastive learning of medical visual representations from paired images and text},
  author={Zhang, Yuhao and Jiang, Hang and Miura, Yasuhide and Manning, Christopher D and Langlotz, Curtis P},
  booktitle={Machine Learning for Healthcare Conference},
  pages={2--25},
  year={2022},
  organization={PMLR}
}

@article{wang2022multi,
  title={Multi-granularity cross-modal alignment for generalized medical visual representation learning},
  author={Wang, Fuying and Zhou, Yuyin and Wang, Shujun and Vardhanabhuti, Varut and Yu, Lequan},
  journal={Advances in neural information processing systems},
  volume={35},
  pages={33536--33549},
  year={2022}
}

@inproceedings{
    yang2025one,
    title={One Leaf Reveals the Season: Occlusion-Based Contrastive Learning with Semantic-Aware Views for Efficient Visual Representation},
    author={Yang, Xiaoyu and Xu, Lijian and Li, Hongsheng and Zhang, Shaoting},
    booktitle={Forty-second International Conference on Machine Learning},
    year={2025}
}

@article{young2026fewer,
  title={Fewer Tokens, Greater Scaling: Self-Adaptive Visual Bases for Efficient and Expansive Representation Learning},
  author={Young, Shawn and Zeng, Xingyu and Xu, Lijian},
  journal={arXiv preprint arXiv:2511.19515},
  year={2026}
}

@article{young2026scalar,
  title={SCALAR: Spatial-Concept Alignment for Robust Vision in Harsh Open World},
  author={Yang, Xiaoyu and Xu, Lijian and Zeng, Xingyu and Wang, Xiaosong and Li, Hongsheng and Zhang, Shaoting},
  journal={Pattern Recognition},
  year={2026}
}

@article{xu2023learning,
  title={Learning a multi-task transformer via unified and customized instruction tuning for chest radiograph interpretation},
  author={Xu, Lijian and Ni, Ziyu and Liu, Xinglong and Wang, Xiaosong and Li, Hongsheng and Zhang, Shaoting},
  journal={arXiv preprint arXiv:2311.01092},
  year={2023}
}

@article{yang2024segmentation,
  title={Segmentation and vascular vectorization for coronary artery by geometry-based cascaded neural network},
  author={Yang, Xiaoyu and Xu, Lijian and Yu, Simon and Xia, Qing and Li, Hongsheng and Zhang, Shaoting},
  journal={IEEE Transactions on Medical Imaging},
  volume={44},
  number={1},
  pages={259--269},
  year={2024},
  publisher={IEEE}
}

@inproceedings{irvin2019chexpert,
  title={Chexpert: A large chest radiograph dataset with uncertainty labels and expert comparison},
  author={Irvin, Jeremy and Rajpurkar, Pranav and Ko, Michael and Yu, Yifan and Ciurea-Ilcus, Silviana and Chute, Chris and Marklund, Henrik and Haghgoo, Behzad and Ball, Robyn and Shpanskaya, Katie and others},
  booktitle={Proceedings of the AAAI conference on artificial intelligence},
  volume={33},
  pages={590--597},
  year={2019}
}

@inproceedings{wang2017chestx,
  title={Chestx-ray8: Hospital-scale chest x-ray database and benchmarks on weakly-supervised classification and localization of common thorax diseases},
  author={Wang, Xiaosong and Peng, Yifan and Lu, Le and Lu, Zhiyong and Bagheri, Mohammadhadi and Summers, Ronald M},
  booktitle={Proceedings of the IEEE conference on computer vision and pattern recognition},
  pages={2097--2106},
  year={2017}
}

@article{shih2019augmenting,
  title={Augmenting the national institutes of health chest radiograph dataset with expert annotations of possible pneumonia},
  author={Shih, George and Wu, Carol C and Halabi, Safwan S and Kohli, Marc D and Prevedello, Luciano M and Cook, Tessa S and Sharma, Arjun and Amorosa, Judith K and Arteaga, Veronica and Galperin-Aizenberg, Maya and others},
  journal={Radiology: Artificial Intelligence},
  volume={1},
  number={1},
  pages={e180041},
  year={2019},
  publisher={Radiological Society of North America}
}

@article{wu2025large,
  title={Large-scale 3d medical image pre-training with geometric context priors},
  author={Wu, Linshan and Zhuang, Jiaxin and Chen, Hao},
  journal={IEEE Transactions on Pattern Analysis and Machine Intelligence},
  year={2025},
  publisher={IEEE}
}

@inproceedings{chen2022multi,
  title={Multi-modal Masked Autoencoders for Medical Vision-and-Language Pre-training},
  author={Chen, Zhihong and Du, Yuhao and Hu, Jinpeng and Liu, Yang and Li, Guanbin and Wan, Xiang and Chang, Tsung-Hui},
  booktitle={International Conference on Medical Image Computing and Computer-Assisted Intervention},
  pages={679--689},
  year={2022},
  organization={Springer}
}

@article{chen2019self,
  title={Self-supervised learning for medical image analysis using image context restoration},
  author={Chen, Liang and Bentley, Paul and Mori, Kensaku and Misawa, Kazunari and Fujiwara, Michitaka and Rueckert, Daniel},
  journal={Medical image analysis},
  volume={58},
  pages={101539},
  year={2019},
  publisher={Elsevier}
}

@inproceedings{kang2023label,
  title={Label-assemble: Leveraging multiple datasets with partial labels},
  author={Kang, Mintong and Li, Bowen and Zhu, Zengle and Lu, Yongyi and Fishman, Elliot K and Yuille, Alan and Zhou, Zongwei},
  booktitle={2023 IEEE 20th International Symposium on Biomedical Imaging (ISBI)},
  pages={1--5},
  year={2023},
  organization={IEEE}
}

@inproceedings{chen2020simple,
  title={A simple framework for contrastive learning of visual representations},
  author={Chen, Ting and Kornblith, Simon and Norouzi, Mohammad and Hinton, Geoffrey},
  booktitle={International conference on machine learning},
  pages={1597--1607},
  year={2020},
  organization={PmLR}
}

@inproceedings{boecking2022making,
  title={Making the most of text semantics to improve biomedical vision--language processing},
  author={Boecking, Benedikt and Usuyama, Naoto and Bannur, Shruthi and Castro, Daniel C and Schwaighofer, Anton and Hyland, Stephanie and Wetscherek, Maria and Naumann, Tristan and Nori, Aditya and Alvarez-Valle, Javier and others},
  booktitle={European conference on computer vision},
  pages={1--21},
  year={2022},
  organization={Springer}
}

@inproceedings{liu2023m,
  title={M-flag: Medical vision-language pre-training with frozen language models and latent space geometry optimization},
  author={Liu, Che and Cheng, Sibo and Chen, Chen and Qiao, Mengyun and Zhang, Weitong and Shah, Anand and Bai, Wenjia and Arcucci, Rossella},
  booktitle={International conference on medical image computing and computer-assisted intervention},
  pages={637--647},
  year={2023},
  organization={Springer}
}

@article{huang2024enhancing,
  title={Enhancing representation in radiography-reports foundation model: a granular alignment algorithm using masked contrastive learning},
  author={Huang, Weijian and Li, Cheng and Zhou, Hong-Yu and Yang, Hao and Liu, Jiarun and Liang, Yong and Zheng, Hairong and Zhang, Shaoting and Wang, Shanshan},
  journal={Nature Communications},
  volume={15},
  number={1},
  pages={7620},
  year={2024},
  publisher={Nature Publishing Group UK London}
}

@inproceedings{zhou2020comparing,
  title={Comparing to learn: Surpassing imagenet pretraining on radiographs by comparing image representations},
  author={Zhou, Hong-Yu and Yu, Shuang and Bian, Cheng and Hu, Yifan and Ma, Kai and Zheng, Yefeng},
  booktitle={International Conference on Medical Image Computing and Computer-Assisted Intervention},
  pages={398--407},
  year={2020},
  organization={Springer}
}

@inproceedings{yue2025chexworld,
  title={Chexworld: Exploring image world modeling for radiograph representation learning},
  author={Yue, Yang and Wang, Yulin and Tao, Chenxin and Liu, Pan and Song, Shiji and Huang, Gao},
  booktitle={Proceedings of the Computer Vision and Pattern Recognition Conference},
  pages={20778--20788},
  year={2025}
}

@article{he2026autoselect,
  title={The Model Knows Which Tokens Matter:Automatic Token Selection via Noise Gating},
  author={He, Landi and Yang, Xiaoyu and Xu, Lijian},
  journal={arXiv preprint arXiv:2603.07135},
  year={2026}
}

@article{chen2026tc,
  title={TC-SSA: Token Compression via Semantic Slot Aggregation for Gigapixel Pathology Reasoning},
  author={Chen, Zhuo and Young, Shawn and Xu, Lijian},
  journal={arXiv preprint arXiv:2603.01143},
  year={2026}
}

@article{xu2024medvilam,
  title={MedViLaM: A multimodal large language model with advanced generalizability and explainability for medical data understanding and generation},
  author={Xu, Lijian and Sun, Hao and Ni, Ziyu and Li, Hongsheng and Zhang, Shaoting},
  journal={arXiv preprint arXiv:2409.19684},
  year={2024}
}

@article{xu2024foundation,
  title={A foundation model for generalizable disease diagnosis in chest X-ray images},
  author={Xu, Lijian and Ni, Ziyu and Sun, Hao and Li, Hongsheng and Zhang, Shaoting},
  journal={arXiv preprint arXiv:2410.08861},
  year={2024}
}

@article{gao2026zerosense,
  title={ZeroSense: How Vision matters in Long Context Compression},
  author={Gao, Yonghan and Chen, Zehong and Xu, Lijian and Chen, Jingzhi and Guan, Jingwei and Zeng, Xingyu},
  journal={arXiv preprint arXiv:2603.11846},
  year={2026}
}

@inproceedings{yang2023geometry,
  title={Geometry-based end-to-end segmentation of coronary artery in computed tomography angiography},
  author={Yang, Xiaoyu and Xu, Lijian and Yu, Simon and Xia, Qing and Li, Hongsheng and Zhang, Shaoting},
  booktitle={International Workshop on Trustworthy Machine Learning for Healthcare},
  pages={190--196},
  year={2023},
  organization={Springer}
}

@article{wu2026multimodal,
  title={Multimodal Model for Computational Pathology: Representation Learning and Image Compression},
  author={Wu, Peihang and Chen, Zehong and Xu, Lijian},
  journal={arXiv preprint arXiv:2603.18660},
  year={2026}
}

@article{chen2026multimodal,
  title={From Snapshots to Symphonies: The Evolution of Protein Prediction from Static Structures to Generative Dynamics and Multimodal Interactions},
  author={Chen, Jingzhi and Xu, Lijian},
  journal={arXiv preprint arXiv:2603.18505},
  year={2026}
}

\end{document}